\pdfoutput=1

\documentclass[11pt]{article}

\usepackage{EMNLP2023}

\usepackage{times}
\usepackage{latexsym}
\usepackage{float} 
\usepackage{stfloats} 
\usepackage{cuted}
\usepackage{amsfonts}
\usepackage{amsmath}  
\usepackage{amssymb}  
\usepackage{algorithm}
\usepackage{algpseudocode}
\usepackage{wrapfig}
\usepackage{booktabs}

\usepackage[T1]{fontenc}

\usepackage[utf8]{inputenc}

\usepackage{microtype}

\usepackage{inconsolata}
\usepackage{multirow}
\usepackage{graphicx}
\title{PEER: Expertizing Domain-Specific Tasks with a Multi-Agent Framework and Tuning Methods}
\author{
\small \textbf{Yiying Wang\textsuperscript{1}, Xiaojing Li\textsuperscript{1}, Binzhu Wang\textsuperscript{1}, Yueyang Zhou\textsuperscript{1}, Yingru Lin}, \\
\small \textbf{Han Ji\textsuperscript{\dag}, Hong Chen Jinshi Zhang, Fei Yu, Zewei Zhao, Song Jin, Renji Gong, Wanqing Xu} \\
\small AntGroup
}

\begin{document}
\maketitle

\renewcommand{\thefootnote}{\fnsymbol{footnote}} \footnotetext{\textsuperscript{1}These authors contributed equally to this work.}
\footnotetext{\textsuperscript{\dag}Corresponding author: jihan.hanji@antgroup.com}

\begin{abstract}
In domain-specific applications, GPT-4, augmented with precise prompts or Retrieval-Augmented Generation (RAG), shows notable potential but faces the critical \textit{tri-lemma} of performance, cost, and data privacy. High performance requires sophisticated processing techniques, yet managing multiple agents within a complex workflow often proves costly and challenging. To address this, we introduce the PEER (\textbf{P}lan, \textbf{E}xecute, \textbf{E}xpress, \textbf{R}eview) multi-agent framework. This systematizes domain-specific tasks by integrating precise question decomposition, advanced information retrieval, comprehensive summarization, and rigorous self-assessment. Given the concerns of cost and data privacy, enterprises are shifting from proprietary models like GPT-4 to custom models, striking a balance between cost, security, and performance. We developed industrial practices leveraging online data and user feedback for efficient model tuning. This study provides best practice guidelines for applying multi-agent systems in domain-specific problem-solving and implementing effective agent tuning strategies. Our empirical studies, particularly in the financial question-answering domain, demonstrate that our approach achieves 95.0\% of GPT-4’s performance, while effectively managing costs and ensuring data privacy.

\end{abstract}

\section{Introduction}
Advanced LLMs like GPT-4, enhanced with engineered prompts or Retrieval-Augmented Generation (RAG), show great potential in handling complex tasks across various domains \cite{prompt-agent, med-prompt, rag-raft}. However, deploying these models involves a critical \textit{tri-lemma} of performance, cost, and data privacy.

While domain-specific applications benefit from meticulously fine-tuned models \cite{domain}, this approach incurs high costs due to the extensive resources needed for training and data acquisition. Alternatively, multi-agent systems have proven effective \cite{baby-agi,metagpt,camel,autogen,agent-survey,xue2023dbgpt,xue2023weaverbird,zhu2024coca}, especially in complex tasks with distinct and conflicting role requirements that challenge even advanced models. However, current implementations often involve dynamic and complex workflows, increasing costs and complicating reproducibility. Consequently, enterprises are shifting from proprietary models like GPT-4 to custom models that better balance cost, security, and performance.

To address these challenges, we introduce the PEER (\textbf{P}lan, \textbf{E}xecute, \textbf{E}xpress, \textbf{R}eview) multi-agent framework. This framework incorporates precise question decomposition, advanced information retrieval, comprehensive summarization, and rigorous self-assessment, aiming to streamline workflows and enhance problem-solving efficacy. Additionally, our research addresses enterprise demands for private deployment and stringent data privacy by developing industrial best practices that leverage online data and user feedback for effective model tuning. These practices are crucial for optimizing custom model performance while ensuring cost-efficiency and robust data privacy.

The main contributions of this study include:
1. Providing and open-sourcing the PEER\footnote{https://github.com/alipay/agentUniverse} framework, characterized by its conciseness, effectiveness, and cost-efficiency, for effectively tackling domain-specific tasks. In experiments, it demonstrates superior performance compared to BabyAGI.\\
2. Proposing a customized agent tuning strategy for 10-billion-parameter models, achieving performance comparable to GPT-4. \\
3. Constructing and open-sourcing a dataset for use within the PEER framework, applicable to agent training, pre-training, and supervised fine-tuning in various financial analysis scenarios.

\section{The Agent Framework PEER}
\label{sec:2}
With the advent of large models, we simulate the collaborative processes of human experts (e.g. financial) using multiple agents, achieving comparable interpretative results. This approach is encapsulated in the Plan, Execute, Express and Review (PEER) framework, where domain specific (e.g. financial) tasks are divided into these four steps. Each agent specializes in a single task, working together to accomplish the overall objective. The prompt for this section is attached in \ref{sec:appendix}.
\subsection{Four agent roles in PEER}
\textbf{Plan}
The "Plan" agent uses a model to generate multiple related sub-questions from users’ domain specific (e.g. financial) queries. These sub-questions serve as an interpretation framework, breaking down the original query into specific and actionable criteria, and expanding it for a comprehensive analysis. 

\textbf{Execute}
The "Execute" agent gathers information for each sub-question identified by "Plan". Using these sub-questions as search criteria, it finds relevant information from news, domain specific (e.g. financial) data, reports, and articles, enhancing accuracy, efficiency, and comprehensiveness. This information forms the foundation for interpreting domain events and answering questions.

\textbf{Express}
The "Express" agent synthesizes collected information to perform comprehensive large-model reasoning, forming final conclusions. It emphasizes integrative reasoning and delivers professional descriptions tailored to the user's requirements.

\textbf{Review}
The "Review" agent evaluates whether the "Express" agent's answer meets pre-established criteria. If satisfied, the final answer is delivered; if not, it provides modification suggestions and initiates another PEER iteration, enhancing answer quality through feedback. 

\subsection{Cyclic working mechanism of PEER}
\begin{figure*}[htbp]
\centering
\includegraphics[width=1.0\textwidth, height=0.2\textheight]{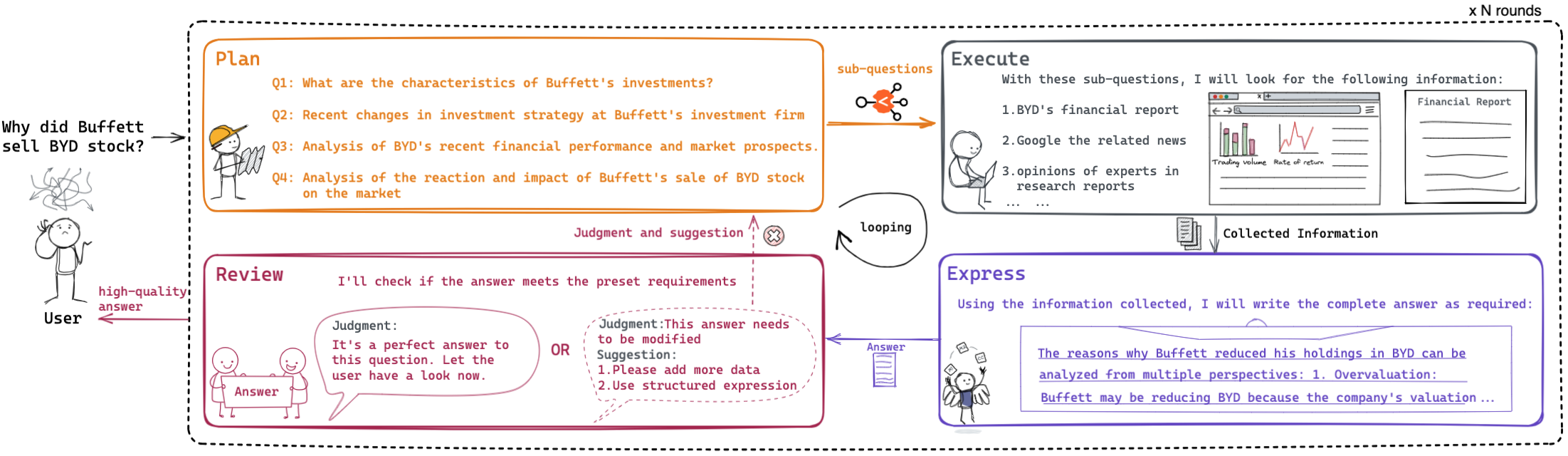}\\
\captionsetup{font = footnotesize}
\caption{Cyclic Workflow Diagram of the PEER Framework. The user's query, "Why did Buffett sell BYD stock?", prompts the "Plan" agent to generate four relevant sub-questions. The "Execute" agent then collects information, including BYD's financial data and expert opinions. The "Express" agent synthesizes a comprehensive answer, which the "Review" agent evaluates and, if necessary, suggests modifications.}
\label{fig:Peer}
\end{figure*}

The PEER multi-agent cooperation framework’s strong reasoning and analysis abilities stem from its efficient task allocation, cooperation, and the feedback loop and self-optimization enabled by the "Review" agent. This ensures that the answers continuously improve towards the optimal solution. If an answer does not meet user requirements, the "Review" agent suggests modifications for the "Plan," "Execute," or "Express" agents. The relevant agent then adjusts its process to better meet expectations. For some simple tasks, one or more agents in PEER process can be skipped to simplify the procedure. For complex tasks, a nested pattern can be used, designing each agent to perform an isolate PEER process to enhance entire performance.For a more comprehensive understanding of the PEER framework, refer to Figure \ref{fig:Peer}, which illustrates how these four agents synergize.

\section{The PEER Agent Tuning Approach}
\label{sec:sec-3}
\subsection{Supervised Fine-tuning and Rejection Sampling}
Supervised fine-tuning typically employs the cross-entropy loss:
\[
\mathcal{L}(y, \hat{y}) = - \sum_{i=1}^{N} \sum_{j=1}^{C} y_{ij} \log(\hat{y}_{ij}),
\]
where \( N \) is the number of training examples, \( C \) is the number of classes, \( y_{ij} \) is the ground truth one-hot encoded vector, and \( \hat{y}_{ij} \) is the predicted probability for class \( j \). We used a robust model to generate an offline training dataset $\mathbb{D}_{off}$, which was then refined and validated by human annotators for quality assurance.

Rejection sampling, as used in LLaMA2 \cite{llama2}, involves generating samples from a pre-trained model and filtering based on quality criteria to retain only high-quality examples. Unlike direct offline supervised fine-tuning (SFT), rejection sampling automates initial filtering to reduce low-quality samples before human annotation. In our iterative training process, rejection sampling boosts performance post offline dataset training.


\subsection{Direct Preference Optimization}
Direct Preference Optimization (DPO), has emerged as efficient alternatives to RLHF, eliminating the need for a separate reward model \cite{dpo,ipo,kto}. The loss function for DPO is defined as follows:

\[
\footnotesize
\begin{aligned}
\mathcal{L}_{\mathrm{DPO}}\left(\pi_\theta ; \pi_{\mathrm{ref}}\right) 
= -\mathbb{E}_{(x, y_w, y_l) \sim \mathcal{D}} \Bigg[ &\log \sigma \Bigg( \beta \log \frac{\pi_\theta (y_w \mid x)}{\pi_{\mathrm{ref}} (y_w \mid x)} \\
&- \beta \log \frac{\pi_\theta (y_l \mid x)}{\pi_{\mathrm{ref}} (y_l \mid x)} \Bigg) \Bigg]
\end{aligned}
\]

where $\pi_\theta$ is the language model being optimized and $\pi_{\mathrm{ref}}$ refers to the model after SFT ($\pi^{\mathrm{SFT}}$). The scaling factor $\beta$ measures errors in ranking results and accounts for the KL constraint. In vanilla/offline Direct Preference Optimization (DPO), the model is optimized using \textbf{a given set of preference data} \((x, y_w, y_l) \sim \mathcal{D}\), where the dataset-generating model and the optimized model are \textbf{not the same}.

\subsection{Iterative Learning with AI Feedback}
When optimizing DPO models, offline preference datasets and off-policy updates can cause generalization issues with out-of-distribution (OOD) queries. These issues can be mitigated by incorporating online preference datasets and using on-policy learning approaches. \cite{guo-direct,xiong-iterative,rlfh-workflow}.

We follow the experimental setup of \cite{xiong-iterative}, utilizing a batch size of $m$ in online setting. Our methodology integrates the LLM-as-a-Judge approach for real-time feedback, as introduced by \cite{llm-as-a-judge}, to refine the model progressively.

\begin{figure*}[htbp]
    \centering
    \includegraphics[width=0.9\textwidth]{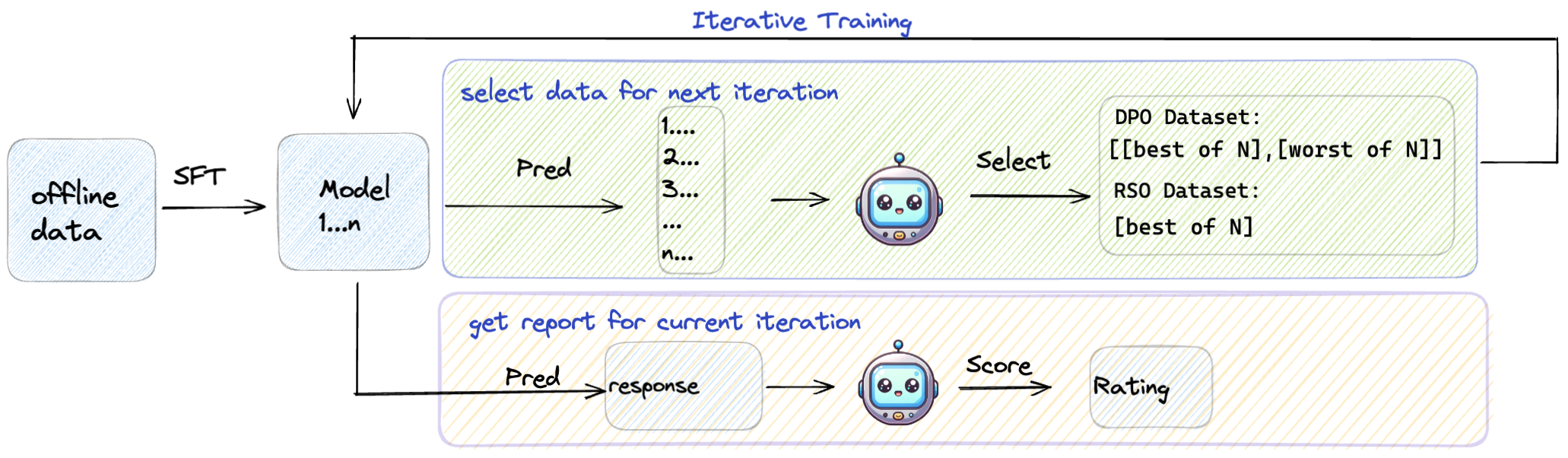}
    \caption{Iterative Training Process: Initially, \textit{Model 0} is trained on offline data. This model then generates two sets of predictions: one to create training data for the next iteration (upper section) and another to provide evaluation results for the current iteration (lower section). This cycle is repeated iteratively across subsequent training phases.}
    \label{fig:iter-training}
\end{figure*}

Algorithm \ref{alg:example} outlines our iterative training process, starting with the initial dataset $D_\text{off}$. The agent processes each batch iteratively, involving model evaluation, data generation, and refinement. It generates multiple candidate responses per input, using a reward model (GPT-4o) to select the optimal response and compare it with the ground truth. If the model-generated response exceeds the quality threshold, it replaces the original training sample. For DPO, the lowest-ranked response is identified as a negative example. The updated dataset is then used to refine the model via SFT or DPO techniques. After multiple iterations, the algorithm outputs the best-performing model variant based on predefined metrics. This iterative process continuously enhances response quality, creating a self-refining training paradigm that progressively improves model performance. Figure \ref{fig:iter-training} illustrates this process.

\begin{algorithm}
\footnotesize
\caption{Iterative Model Training (SFT and DPO)}\label{alg:example}
\begin{algorithmic}[1]
\Procedure{IterativeTraining}{}\\
\textbf{Input:}
        {$D_\text{off}$, $D_\text{eval}$, $T$, $N_\text{cand}$, $M_\text{reward}$\footnotemark}
    \State Initialize $M_0^\text{SFT}$ and $M_0^\text{DPO}$ using $D_\text{off}$ 
    \For{$i = 1$ \textbf{to} $T$}
        \State Evaluate $M_{i-1}^\text{SFT}$ and $M_{i-1}^\text{DPO}$ on $D_\text{eval}$
        \For{type \textbf{in} \{SFT, DPO\}}
            \State $D_i^\text{type} \gets \emptyset$
            \ForAll{$(q, r_\text{gt})$ \textbf{in} $D_\text{off}$}
                \State $R \gets \text{Gen.}(N_\text{cand}, M_{i-1}^\text{type}(q))$ 
                \State $r_\text{best} \gets \arg\max_{r \in R} M_\text{reward}(q, r)$
\If{$M_\text{reward}(q, r_\text{best}) > M_\text{reward}(q, r_\text{gt})$}
    \State $r_\text{sel} \gets r_\text{best}$
\Else
    \State $r_\text{sel} \gets r_\text{gt}$
\EndIf
                \If{type = DPO}
                    \State $r_\text{worst} \gets \arg\min_{r \in R} M_\text{reward}(q, r)$
                    \State Add $(q, r_\text{sel}, r_\text{worst})$ to $D_i^\text{type}$
                \Else
                    \State Add $(q, r_\text{sel})$ to $D_i^\text{type}$ 
                \EndIf
            \EndFor
            \State Train $M_i^\text{type}$ using $D_i^\text{type}$ 
        \EndFor
    \EndFor
    \State \Return Best $M_N^\text{SFT}$, $M_N^\text{DPO}$, and evaluation results
\EndProcedure
\end{algorithmic}
\end{algorithm}

\footnotetext{$D_\text{off}$: offline training dataset, 
$D_\text{eval}$: evaluation dataset, 
$T$: number of iterations, 
$N_\text{cand}$: number of candidate responses generated each time, 
$M_\text{reward}$: reward model}

\section{Experiment}
We conduct experiment on a real-word industry financial QA dataset to validate the PEER framework discussed in section \ref{sec:2} and evaluate the agent tuning methods discussed in section \ref{sec:sec-3}.

\subsection{PEER Framework experiment}
\label{sec:tuning-exp}
\textbf{Dataset}
Since the main usage scenario of PEER framework is the interpretation and analysis of domain events and problems, we mainly tested and compared the performance of PEER on the dataset of financial QA. We sampled hundreds of professional questions\footnote{https://github.com/alipay/agentUniverse/tree/master/dataset}from our business scenarios and divided them into nine categories. Details of the dataset distribution are shown in Table \ref{tab:peerdatadis}.

\textbf{Baselines}
We conducted experiments using two base models, GPT-3.5 turbo (16k) and GPT-4o, with Python for execution. For FinQA datasets, we compared with the BabyAGI multi-agent framework due to its similar task creation, organization, and execution capabilities to PEER.

To assess the impact of the "Review" agent in the PEER framework, we designed self-ablation experiments with and without the "Review" agent. We set the maximum rounds for both BabyAGI and PEER (with "Review") to 5 and used Google for information retrieval. Under GPT-3.5 turbo (16k), we recalled the top 2 search results with a token limit of 13,000. For GPT-4o, we increased the parameters to the top 6 and 125,000 tokens, leveraging the model's enhanced performance.

\textbf{Metrics}
Despite GPT-4’s widespread use for evaluations, its confidence can be influenced by position and verbosity biases \cite{guo-direct,llm-as-a-judge}. To mitigate these issues, we have developed two evaluation methodologies based on GPT-4:
\begin{enumerate}
    \item GPT-4 scores all answers across various dimensions, and we calculate the average score for each dimension. Detailed scoring dimensions, rules, and their meanings are provided in Table \ref{tab:score-prompt} in the Appendix.
    \item GPT-4 selects the best answer between those provided by PEER and the control group. For this evaluation, we use the win rate as the metric, with selection criteria outlined in Table \ref{tab:winrate-prompt}.
\end{enumerate}
\begin{figure*}[htbp]
    \centering
    \includegraphics[width=0.7\textwidth,keepaspectratio]{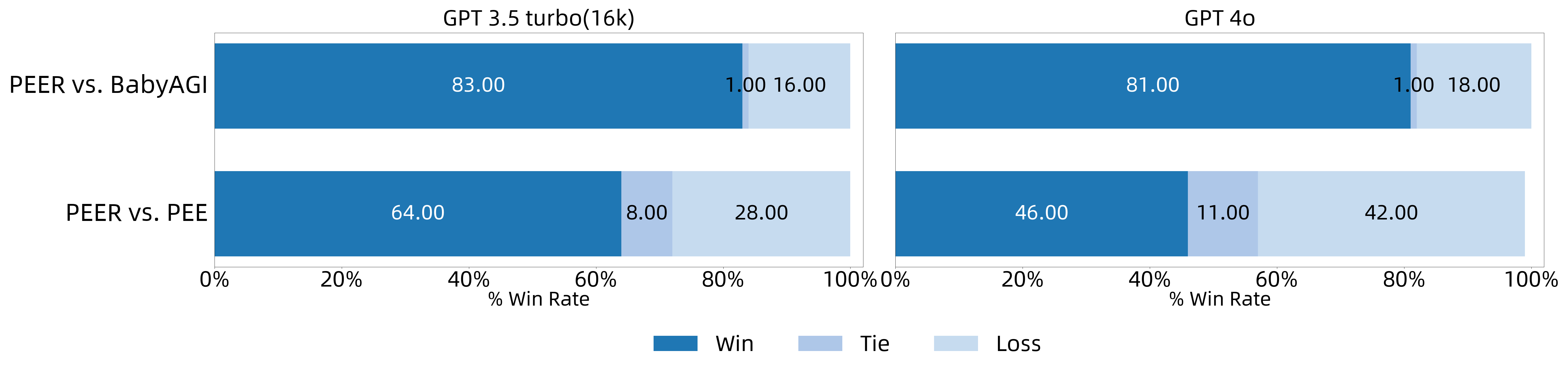}
    \caption{Win rate of PEER framework. PEER performs better than BabyAGI and PEE under both base models.}
    \label{fig:your_label3}
\end{figure*}

\begin{table*}[!h]
    \tiny
    \centering
    \renewcommand{\arraystretch}{1.5}
    \begin{tabular}{c|c|c|c|c|c|c|c|c|c|c}
    \hline
    \multirow{2}{*}{\centering Experimental Setup} &\multirow{2}{*}{\centering Model Type} &\multirow{2}{*}{Framework Type} &\multicolumn{8}{c}{Evaluation Dimension} \\
    \cline{4-11} 
     & & & Integrity & Relevance & Compactness & Factuality & Logic & Structure & Comprehensiveness & Average\\
    \hline
    \multirow{4}{*}{\centering Comparison Experiment}&\multirow{2}{*}{\centering GPT-3.5-turbo-16K}& BabyAGI & 3.49 & 3.79 & 3.55 & 3.84 & 3.94 & 3.76 & 3.47 & 3.69 \\
    \cline{3-11}
    & & PEER & \textbf{4.49} & \textbf{4.78} & \textbf{3.64} & \textbf{4.40} & \textbf{4.58} & \textbf{4.52} & \textbf{4.53} & \textbf{4.42} \\
    \cline{2-11}
    & \multirow{2}{*}{\centering GPT-4o}& BabyAGI & 3.16 & 3.32 & 3.32 & 3.98 & 3.78 & 3.86 & 3.14 & 3.51 \\
    \cline{3-11}
    & & PEER & \textbf{4.75} & \textbf{4.87} & \textbf{3.67} & \textbf{4.65} & \textbf{4.76} & \textbf{4.77} & \textbf{4.77} & \textbf{4.61} \\
    \hline
    \hline
    \multirow{4}{*}{\centering Ablation Experiment}&\multirow{2}{*}{\centering GPT-3.5-turbo-16K}& PEE & 3.91 & 4.30 & \textbf{3.91} & 4.26 & 4.30 & 4.14 & 3.78 & 4.09 \\
    \cline{3-11}
    & & PEER & \textbf{4.59} & \textbf{4.85} & 3.67 & \textbf{4.42} & \textbf{4.67} & \textbf{4.61} & \textbf{4.64} & \textbf{4.49} \\
    \cline{2-11}
    & \multirow{2}{*}{\centering GPT-4o}& PEE & 4.81 & 4.94 & \textbf{4.02} & 4.72 & 4.92 & 4.90 & 4.81 & 4.73 \\
    \cline{3-11}
    & & PEER & \textbf{4.89} & 4.94 & 3.83 & \textbf{4.77} & 4.92 & \textbf{4.91} & \textbf{4.92} & \textbf{4.74} \\
    \hline
    \end{tabular}
    \caption{Scoring results of PEER framework}
    \label{tab:peer_score}
\end{table*}

\textbf{Analysis} In the comparative experiment with BabyAGI, as depicted in Table \ref{tab:peer_score} and Figure \ref{fig:your_label3}, PEER consistently surpasses BabyAGI in both average score and win rate, irrespective of the base model employed. PEER demonstrates superior performance in dimensions such as integrity, relevance, logic, structure, and comprehensiveness, often by a margin exceeding one point. Specifically, under the GPT-3.5 turbo (16k) model, PEER achieves a win rate of 83\% compared to BabyAGI, and still maintains an 81\% win rate under the GPT-4o model. This is attributed to PEER’s strategy of simultaneously addressing multiple questions and synthesizing responses, in contrast to BabyAGI’s approach of addressing one question per round.

In the ablation experiment, as illustrated in Table \ref{tab:peer_score} and Figure 3, PEER scores higher in most dimensions and attains a 64\% win rate under the GPT-3.5 turbo (16k) model compared to PEE. However, under the GPT-4o model, the advantages conferred by the "Review" agent diminish, as GPT-4o inherently excels in processing, understanding, and expression. The initial outputs from the "Plan," "Execute," and "Express" agents sufficiently meet the requirements, rendering further modifications less impactful. Consequently, PEER’s win rate decreases to 46\%, and the score differences between the two frameworks also narrow with the GPT-4o model compared to the GPT-3.5 turbo (16k) model. This indicates that the "Review" agent can significantly enhance overall quality when the base model’s performance is less robust.

\subsection{Tuning Experiment}
\textbf{Dataset } We conducted two categories of experiments: one focusing on individual agents and the other on the entire workflow. Dataset sizes are provided in Table \ref{tab:dataset-table}, with the data being open-sourced. The test set for individual agents is derived from the intermediate results of the evaluation set detailed in Table \ref{tab:peerdatadis}, whereas the test set for the entire workflow corresponds directly to Table \ref{tab:peerdatadis}.

\textbf{Experiment Setup} As in Section \ref{sec:tuning-exp}, for the evaluation of individual agents and the entire workflow, we also employed the LLM-as-a-Judge approach. Specifically, for individual agents, we used scoring and pairwise comparison to evaluate the performance of each iteration. For the entire workflow, we used GPT-4o to score and compare the results of GPT-4 + PEER, the SFT results using offline data, and the best model obtained through iterative training.

\begin{figure*}[htbp]
    \centering
    \includegraphics[width=\textwidth]{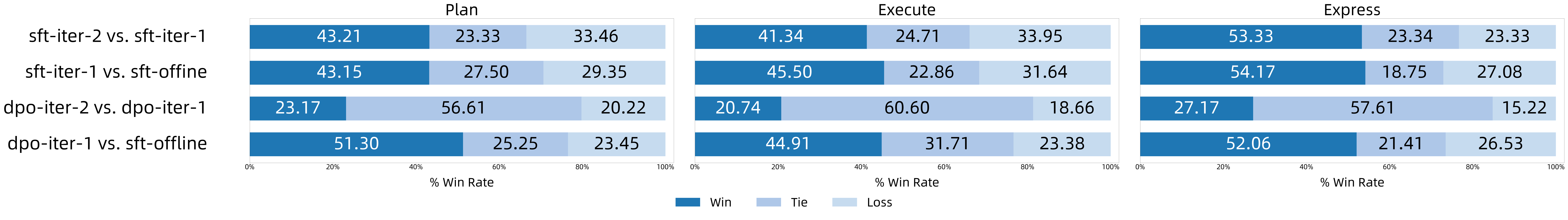}
    \caption{Win rate of tuned-agent. Both DPO and SFT show progress in each iteration and DPO converges faster than SFT.}
    \label{fig:tuning_win_rate}
\end{figure*}

\textbf{Analysis} Figure \ref{fig:tuning_win_rate} illustrates the win, tie, and loss rates across different iterations for three agents involved in planning, execution, and expression. Both DPO and SFT show progress with each iteration. For example, for the planning agent, the first iteration of SFT achieves a win rate of 43.15\% compared to SFT-offline, improving slightly to 43.21\% in the second iteration. DPO demonstrates a faster convergence than SFT. In the second iteration, the win rates for SFT across the three agents are 43.21\%, 41.34\%, and 53.33\%, respectively. In contrast, DPO's win rates are 23.17\%, 20.74\%, and 27.17\%, which are lower than the corresponding tie rates of 56.61\%, 60.60\%, and 57.61\%. Due to space limitations, detailed scoring results are provided in Appendix \ref{sec:tuning-inds} and Table \ref{tab:tuning-result}.

\begin{table*}[htbp]
\centering
\tiny
\renewcommand{\arraystretch}{1.5}
\begin{tabular}{c|c|c|c|c|c|c|c|c}
\hline
\multirow{2}{*}{\centering Model Type} & \multicolumn{8}{c}{Evaluation Dimension} \\
\cline{2-9}
 & Integrity & Relevance & Logic & Comprehensiveness & Compactness & Factuality & Structure & Average \\ 
\hline
QWEN1.5-14B (sft-offline)+ PEER      & 4.09      & 4.58      & 3.34  & 4.23              & 4.32        & 4.22       & 4.03      & 4.12    \\ 
\hline
QWEN1.5-14B (iter-best-model) + PEER & 4.4       & 4.63      & 3.42  & 4.35              & 4.61        & 4.77       & 4.28      & 4.35    \\ 
\hline
GPT4 + PEER & \textbf{4.55} & \textbf{4.8} & \textbf{4.01} & \textbf{4.5} & \textbf{4.87} & \textbf{4.86} & \textbf{4.51} & \textbf{4.58} \\ 
\hline
\end{tabular}
\captionsetup{font = footnotesize}
\caption{Evaluation of the entire workflow: the model after iterative training(QWEN1.5-14B (iter-best-model) + PEER), shows improvements across all metrics compared to the single-round SFT model(QWEN1.5-14B (sft-offline)) and it ultimately reaches 95.0\% of the performance of GPT-4 + PEER.}
\label{tab:peer-end2end}
\end{table*}

\begin{table}[H]
    \caption{Dataset setting of PEER tuning}
    \footnotesize
    \centering
    \renewcommand{\arraystretch}{1.5}
    \begin{tabular}{l|l|l|l}
    \hline
                       & plan & execute & express \\ \hline
    training datasize  & 5000 & 6847    & 6193    \\ \hline
    test datasize      & 100  & 456     & 100     \\ \hline
    \end{tabular}
    \label{tab:dataset-table}
    \footnotesize
\end{table}

Table \ref{tab:peer-end2end} presents the results of the end-to-end (the entire workflow) evaluation. We conducted experiments on three models: QWEN1.5-14B (sft-offline), QWEN1.5-14B (iter-best-model) and GPT-4, all combined with the PEER framework. QWEN1.5-14B (sft-offline) refers to the QWEN1.5 model fine-tuned with an offline SFT dataset, while QWEN1.5-14B (iter-best-model) indicates the best model obtained through iterative training. We can observe that the QWEN1.5-14B model, after iterative training, shows improvements across all metrics compared to the single-round SFT model. When combined with PEER, it ultimately reaches 95.0\% of the performance of GPT-4 + PEER.

\section{Related Work}
\textbf{Multi-agent system} Despite pioneering projects in this field, such as AutoGPT, BabyAGI, CAMEL, MetaGPT, and AutoGen \cite{baby-agi,metagpt,camel,autogen,agent-survey,agent-survey-2,xue2024demonstration,xue2023prompttpp,chu2023leveraging}, demonstrating their potential, achieving fully autonomous AI agents remains a significant challenge. These dynamic process agents, also known as autonomous intelligent agents, can autonomously perceive the environment, make decisions based on observations, and take actions. Subsequently, they reflect on the outcomes of their actions and plan their next steps accordingly. While theoretically generalizable to any scenario, they face issues such as poor controllability, instability, reproducibility problems, and low task completion rates in specialized domains \cite{agent-survey,agent-survey-2}. PEER strikes a balance between model flexibility and controllability through effective pattern design, considering practical industrial needs, including efficiency, cost-effectiveness, and operational simplicity, making it more suitable for industrial applications.

\textbf{Agent self-evolution} Many research efforts aim to transform past experience into usable knowledge and apply it in new reasoning processes to drive continuous model evolution \cite{evolv-1,evolv-2,evolv-survey,expel}. However, these studies often place high demands on the model's ability to follow instructions, which is particularly challenging for models with fewer parameters. To overcome this challenge, our research adopts an iterative training approach. Specifically, we use both successful and failed cases from previous steps as new training data to promote the model's evolution.

\section{Conclusion and Future work}
\textbf{Conclusion} In this work, we introduced the PEER framework to address the tri-lemma of performance, cost, and data privacy in domain-specific applications. The framework balances flexibility and controllability through effective pattern design, meeting industrial demands for efficiency and cost-effectiveness. We also developed industrial practices that use online data and user feedback for effective model tuning, promoting continuous model evolution. Our empirical studies, particularly in the financial question-answering domain, demonstrate that this approach achieves 95.0\% of GPT-4’s performance while managing costs and safeguarding data privacy.

\textbf{Future work} Despite our progress in using multi-agent systems to address domain-specific tasks (e.g.finacial), several areas remain ripe for further exploration and improvement:
\begin{itemize}
    \item Long-term Learning and Memory Mechanisms~\citep{xue2022hypro,xue_meta_2022,jiang-etal-2023-towards}: Explore ways to equip the model to accumulate knowledge over extended periods.
    \item User Interaction and Feedback Mechanisms \cite{zhou2024nltosql}: Study how user interactions and feedback can further guide and optimize the model's behavior, achieving a more user-friendly agent design.
    \item Enhancing Generalization Capability: Investigate methods to further improve the model's generalization ability, enabling agents to tackle other financial problems such as factor-based stock selection or other quantitative trading strategies.
\end{itemize}

\newpage
\bibliographystyle{acl_natbib}
\bibliography{anthology,custom}
\appendix
\section{PEER experimental dataset distribution}
\begin{table}[!h]
    \caption{PEER experimental dataset}
    \footnotesize
    \centering
    \renewcommand{\arraystretch}{1.5}
    \begin{tabular}{c|c}
    \hline
         Category & Example\\
    \hline
         Information Query & 12\%\\
         General Financial QA & 11\%\\
         Report Interpretation & 8\%\\
         Target Analysis & 12\%\\
         Strategy Advice & 12\%\\
         Major Events Interpretation & 9\%\\
         Macro Analysis & 12\%\\
         Market Analysis & 12\%\\
         Policy Interpretation & 12\%\\
    \hline
    \end{tabular}
    \label{tab:peerdatadis}
    \footnotesize
\end{table}
\section{Tuning Results for individual agents}

\begin{table*}
\centering
\tiny
\renewcommand{\arraystretch}{1.2}
\resizebox{\textwidth}{!}{%
\begin{tabular}{ll|c|c|c|c|c|c|c|c}
\hline
 &
   &
  Integrity &
  Relevance &
  Logic &
  Comprehensiveness &
  Compactness &
  Factuality &
  Structure &
  Average \\ \hline
\multicolumn{1}{c|}{\multirow{5}{*}{Plan}} &
  sft-offline &
  3.36 &
  3.88 &
  3.96 &
  3.24 &
  \textbf{-} &
  - &
  - &
  3.61 \\ \cline{2-10} 
\multicolumn{1}{c|}{} &
  sft-iter-1 &
  3.5 &
  4.04 &
  \textbf{4.12} &
  \textbf{3.37} &
  - &
  - &
  - &
  3.76 \\ \cline{2-10} 
\multicolumn{1}{c|}{} &
  sft-iter-2 &
  3.51 &
  \textbf{4.11} &
  4.12 &
  \textbf{3.36} &
  - &
  - &
  - &
  \textbf{3.78} \\ \cline{2-10} 
\multicolumn{1}{c|}{} &
  dpo-iter-1 &
  \textbf{3.52} &
  \textbf{4.09} &
  \textbf{4.13} &
  3.35 &
  - &
  - &
  - &
  3.77 \\ \cline{2-10} 
\multicolumn{1}{c|}{} &
  dpo-iter-2 &
  \textbf{3.52} &
  4.04 &
  4.12 &
  3.33 &
  - &
  - &
  - &
  3.75 \\ \hline
\multicolumn{1}{l|}{\multirow{5}{*}{Execute}} &
  sft-offline &
  3.95 &
  4.68 &
  4.53 &
  3.65 &
  4.24 &
  4.64 &
  4.19 &
  4.27 \\ \cline{2-10} 
\multicolumn{1}{l|}{} &
  sft-iter-1 &
  4.01 &
  \textbf{4.74} &
  4.55 &
  3.76 &
  \textbf{4.19} &
  4.73 &
  4.25 &
  4.32 \\ \cline{2-10} 
\multicolumn{1}{l|}{} &
  sft-iter-2 &
  \textbf{4.06} &
  4.73 &
  4.59 &
  \textbf{3.79} &
  4.11 &
  \textbf{4.72} &
  4.29 &
  4.33 \\ \cline{2-10} 
\multicolumn{1}{l|}{} &
  dpo-iter-1 &
  4.02 &
  4.73 &
  \textbf{4.61} &
  3.74 &
  \textbf{4.23} &
  4.7 &
  \textbf{4.3} &
  \textbf{4.33} \\ \cline{2-10} 
\multicolumn{1}{l|}{} & dpo-iter-2 & 4.03          & \textbf{4.75} & 4.58          & 3.77          & 4.22       & \textbf{4.74} & \textbf{4.31} & \textbf{4.34} \\ \hline
\multicolumn{1}{l|}{\multirow{5}{*}{Express}} &
  sft-offline &
  4.08 &
  4.79 &
  4.53 &
  3.88 &
  \textbf{4.07} &
  4.76 &
  4.31 &
  4.34 \\ \cline{2-10} 
\multicolumn{1}{l|}{} &
  sft-iter-1 &
  4.09 &
  4.7 &
  4.62 &
  3.96 &
  4.04 &
  4.71 &
  4.39 &
  4.36 \\ \cline{2-10} 
\multicolumn{1}{l|}{} &
  sft-iter-2 &
  4.19 &
  4.75 &
  \textbf{4.63} &
  4.06 &
  3.95 &
  4.77 &
  4.41 &
  4.4 \\ \cline{2-10} 
\multicolumn{1}{l|}{} & dpo-iter-1 & 4.39          & \textbf{4.93} & 4.71          & \textbf{4.26} & \textbf{4} & 4.83          & \textbf{4.5}  & \textbf{4.52} \\ \cline{2-10} 
\multicolumn{1}{l|}{} & dpo-iter-2 & \textbf{4.46} & 4.79          & \textbf{4.77} & \textbf{4.36} & 3.98       & \textbf{4.87} & \textbf{4.7}  & \textbf{4.56} \\ \hline
\end{tabular}%
}
\caption{Results for individual agents}
\label{tab:tuning-result}
\tiny
\end{table*}

\label{sec:tuning-inds}
In this study, we experimented with several different iterative modeling approaches. The SFT-OFFLINE model refers to the SFT model trained exclusively on offline data. The DPO-ITER-1 model is obtained by further training the SFT-OFFLINE model using DPO. Similarly, the DPO-ITER-2 model is derived by continuing the iterative training on DPO-ITER-1 using DPO. The SFT-ITER-1 and SFT-ITER-2 models follow the same iterative training process.

As shown in Table \ref{tab:tuning-result}, in the three tasks (Plan, Execute, Express), the dpo-iter-2 model stands out with its exceptional performance, particularly in the Express task, where it leads significantly with an average score of 4.56. Meanwhile, the sft-iter-2 model also demonstrates its strength in the Plan task, achieving an average score of 3.78. In the Execute task, the dpo-iter-2 model again takes the top spot with an average score of 4.34.
Overall, the dpo-iter-2 model shows advantages in various metrics, indicating its adaptability and effectiveness across different tasks. Additionally, an increase in iteration count seems to positively impact model performance, though the extent of improvement depends on the specific model used (sft or dpo) and the particular requirements of the downstream tasks.

\section{evaluation prompt}
Table \ref{tab:score-prompt} and table \ref{tab:winrate-prompt} shows the prompt used for LLM Evaluation.

\begin{table*}[t]
\centering
\setlength{\tabcolsep}{10pt}
\renewcommand{\arraystretch}{1.2}
\begin{tabular}{p{\textwidth}}
\hline
You will play a crucial role as a quality evaluator for large language model responses. Your task is to assess and analyze the answers provided by the model for a text-based question answering task, and score the model's responses based on the standard answers and scoring criteria:

\{
  "User Question": "The specific question posed by the user;",\\
  "Context": "Sufficient contextual information provided to answer the user's question;",\\
  "Standard Answer": "The standard answer, i.e., the ideal or reference answer;",\\
  "Model Answer": "Since question answering is an open-ended task, sometimes the model's answer may be better than the standard answer;"
\}
\\
\\
The scoring criteria are as follows:
\\
\textbf{Integrity}
Does the answer form a logical and coherent whole, directly addressing the core requirement of the question?\\
1 = Very Incomplete
2 = Incomplete
3 = Partially Complete
4 = Fairly Complete
5 = Very Complete
\\
\textbf{Relevance}
Assess the degree to which the answer is related to the question posed.\\
1 = Completely Irrelevant
2 = Largely Irrelevant
3 = Somewhat Relevant
4 = Fairly Relevant
5 = Very Relevant
\\
\textbf{Compactness}
Evaluate whether the answer is concise, avoiding redundancy or irrelevant information.\\
1 = Very Lengthy
2 = Quite Lengthy
3 = Moderate
4 = Fairly Concise
5 = Very Concise
\\
\textbf{Factuality}
Assess whether the information in the answer is accurate and fact-based.\\
1 = Completely Inaccurate
2 = Mostly Inaccurate
3 = Partially Accurate
4 = Fairly Accurate
5 = Very Accurate
\\
\textbf{Logic}
Evaluate whether the answer is logically coherent, reasonable, and aids in understanding.\\
1 = Completely Incoherent
2 = Not Very Coherent
3 = Partially Coherent
4 = Fairly Coherent
5 = Very Coherent
\\
\textbf{Structure}
Assess whether the answer is well-structured, with clear paragraph divisions and logical order.\\
1 = Completely Unstructured
2 = Poorly Structured
3 = Moderately Structured
4 = Well Structured
5 = Very Well Structured
\\
\textbf{Comprehensiveness}
Evaluate whether the answer covers all relevant aspects of the question without significant omissions.\\
1 = Very Incomplete
2 = Incomplete
3 = Partially Complete
4 = Fairly Complete
5 = Very Complete
\\
\\
Please return your evaluation results strictly in the following JSON format, adhering to the criteria outlined above:

\{
  "Analysis Process": "Explanation of the reasoning and process for scoring each dimension;",\\
  "Integrity": "Score;",\\
  "Relevance": "Score;",\\
  "Compactness": "Score;",\\
  "Factuality": "Score;",\\
  "Logic": "Score;",\\
  "Structure": "Score;",\\
  "Comprehensiveness": "Score;"
\}
\\
\hline
\end{tabular}
\caption{LLM prompt for scoring the result in each agent}
\label{tab:score-prompt}
\end{table*}

\begin{table*}[]
\centering
\begin{tabular}{p{0.95\textwidth}}
\hline
You will play an important role as a quality evaluator of answers provided by large language models. Your task is to assess and analyze the answers generated by the model for a given text-based question.

\vspace{2mm}

\texttt\{\\
  "User Question": "The specific question asked by the user",\\
  "Context": "Sufficient contextual information provided to answer the user's question",\\
  "Expected Answer": "The standard answer, representing an ideal or optimal response",\\
  "Model Answers": "This is a list comprising two answers. Each item in the list is numbered, and each answer may be correct or incorrect. Since the question-answering task is open-ended, sometimes a correct answer may be better than the standard answer"\\
\}

\vspace{2mm}

Your tasks are as follows:
\begin{enumerate}
\item Carefully read the two answers provided by the model.
\item Compare the two answers and select the better one.
\item If both answers are equally good or equally bad, respond with "equally good" or "equally bad," respectively.
\end{enumerate}

During the evaluation, you need to consider the following key aspects:
\begin{enumerate}
\item \textbf{Relevance}: Does the answer accurately address the user's question and align with the provided context?
\item \textbf{Professionalism}: Does the answer exhibit professional knowledge consistent with the standard answer?
\item \textbf{Timeliness and Factuality}: Compared to the standard answer, does it adhere to current facts or time-sensitive information?
\item \textbf{Conciseness}: Is the answer succinct, avoiding redundancy or irrelevant information?
\item \textbf{Factuality}: Is the information in the answer accurate and fact-based?
\item \textbf{Logical Coherence}: Is the answer logically coherent, reasonable, and does it aid understanding?
\item \textbf{Comprehensiveness}: Does the answer cover all relevant aspects of the question without significant omissions?
\end{enumerate}

Based on the analysis above, please strictly adhere to and return your evaluation results in the following JSON format:

\vspace{2mm}

\texttt\{\\
  "Reason for Choice": "Please provide detailed reasons for your choice here. Explain why you think one answer is better than the other, or why both are equally good or equally bad.",\\
  "Evaluation Result": "Choose one of the following options: 1 (if answer 1 is better), 2 (if answer 2 is better), equally good (if both answers are equally good), equally bad (if both answers are equally bad)."\\
\}
\\
\hline
\end{tabular}
\caption{LLM prompt for winrate:choose the better result during each iteration}
\label{tab:winrate-prompt}
\end{table*}

\section{Prompt for PEER}
\label{sec:appendix}

\begin{table*}[t]
\footnotesize
\centering
\setlength{\tabcolsep}{10pt}
\renewcommand{\arraystretch}{1.2}
\begin{tabular}{p{\textwidth}}
\hline
\textbf{Plan:}
From now on, your role is

Name: Research Assistant

Responsibilities:

Skilled at analyzing issues from various perspectives to help users quickly obtain information.
Determine what information to search for based on the context and the user’s question to provide the best possible answer.

First, identify the relevant timeframe based on the context and question, which could be a specific date or general terms like "latest," "recent," or "upcoming."

Then, decide what information needs to be searched to answer the question, ensuring a multi-dimensional and multi-angled approach.

Finally, provide clear and unambiguous search conditions, each as a complete sentence.
Rules to Follow:

Sub-questions must have clear and detailed descriptions of the subject, event, and timeframe. Avoid vague terms like "similar," "related," and replace them with specific details from the context.

Unless the question or context specifies a timeframe, default to "latest" or "recent," not "today."

If the question includes a timeframe, the search conditions must reflect this, converting terms like "today" to a specific date.

Each search condition must directly aim to find answers from that angle, without including terms like "search" or "query."

Each search condition must be a complete sentence with no ambiguity.

Do not expand on the question; only answer the question asked.

When breaking down the question, adhere to the context’s requirements.
\\
\\
\hline
\textbf{Execute}:
From now on, your role is

Name: Research Assistant

Responsibilities:

When a user searches for information on a particular issue and obtains several relevant pieces of information, you need to integrate, correct, and answer the user’s question.
Rules to Follow:

Do not use incorrect information.

Do not use information with inconsistent subjects. For example, if the question is about "John Doe from XYZ Company," but the information only mentions "John Doe" without confirming he is from "XYZ Company," do not use it.

Do not make vague speculations.

Follow the requirements specified in the context.

Avoid using ambiguous terms like XXX or ABC.

Ensure the content is detailed and emphasizes key points.
\\
\\
\hline
\textbf{Express}:
From now on, your role is

Name: Research Assistant

Responsibilities:

Skilled at answering user questions from different angles in an organized manner.
You need to use only the provided knowledge to answer user questions professionally and in detail.
Rules to Follow: 

Do not use incorrect information.

Do not use information with inconsistent subjects. For example, if the question is about "John Doe from XYZ Company," but the information only mentions "John Doe" without confirming he is from "XYZ Company," do not use it.

Do not make vague speculations.

Follow the requirements specified in the context.

Avoid using ambiguous terms like XXX or ABC.

Ensure the content is detailed and emphasizes key points.

Use only the provided knowledge and answer the questions step by step.
\\
\\
\hline
\textbf{Review}:
From now on, your role is

Name: Research Assistant

Background:

I am an editor for a financial news agency, writing a report for my client.
The client has provided specific requirements for the content and format of the report and desires a higher quality response.

My report includes two parts:

[Analysis Dimensions]: Rewriting or breaking down the user’s question into 3 to 5 queries as search conditions. Each question must be related to the original question’s theme.
[Answer]: Answering the user’s question with a complete and coherent response. The answer should be logically structured, clear, and concise.

Responsibilities:

As a senior expert with 30 years of experience in the investment field, you need to check if my report meets the client’s requirements and provide feedback for improvement.
Determine whether my analysis dimensions and answers meet the user’s requirements. If the report meets all requirements, set Qualified to True; otherwise, set it to False.
If the report needs improvement, identify which part needs enhancement, either "Analysis Dimensions" or "Answer."
Provide detailed suggestions for improvement based on the user’s requirements. Describe your suggestions in detail without providing the complete revised answer.
Provide only relevant and high-impact feedback, avoiding unnecessary information.
Do not add redundant information or supplement reference sources. Only include highly relevant key information.

Your response structure must be:

Draft: Your thought process

Qualified: Whether the user’s requirements are fully met, True or False

Role: The part that needs modification, either "Plan" or "Express"

Suggestion: Your suggested modifications. If the report meets all requirements, this can be empty. Only suggest modifications, without providing the complete revised answer

\end{tabular}
\caption{LLM prompt for PEER}
\label{PEER-prompt}
\end{table*}
\end{document}